\pdfoutput=1
\documentclass[11pt]{article}
\usepackage{acl2012}
\makeatletter
\newcommand{\@BIBLABEL}{\@emptybiblabel}
\newcommand{\@emptybiblabel}[1]{}
\makeatother
\usepackage{times}
\usepackage{latexsym}
\usepackage{amsmath}
\usepackage{multirow}
\usepackage{placeins}
\usepackage{url}
\usepackage{array}
\usepackage{multirow}
\usepackage{fancyhdr} 
\usepackage{hyperref}

\title{Statistical Modality Tagging\\
from Rule-based Annotations and Crowdsourcing}

\author{\rm
\begin{tabular}{ccc}
{\bf Vinodkumar Prabhakaran} & {\bf Michael Bloodgood} & {\bf Mona Diab}\\
CS & CASL & CCLS\\
Columbia University & University of Maryland & Columbia University\\
vinod@cs.columbia.edu & meb@umd.edu & mdiab@ccls.columbia.edu\\
\\
\\
{\bf Bonnie Dorr} & {\bf Lori Levin} & {\bf Christine D. Piatko}\\
CS and UMIACS & LTI & APL \\
University of Maryland & Carnegie Mellon University & Johns Hopkins University \\
bonnie@umiacs.umd.edu & lsl@cs.cmu.edu & christine.piatko@jhuapl.edu\\
\end{tabular}
\\
\\
\\
\begin{tabular}{cc}
{\bf Owen Rambow} & {\bf Benjamin Van Durme}\\
CCLS & HLTCOE\\
Columbia University & Johns Hopkins University \\
rambow@ccls.columbia.edu & vandurme@cs.jhu.edu\\
\end{tabular}
}

\date{}

\begin{document}

\thispagestyle{fancy}

\maketitle
\begin{abstract}
We explore training an automatic modality tagger. 
Modality is the attitude that a speaker might have toward an event or state.
One of the main hurdles for training a linguistic tagger is gathering training data.  
This is particularly problematic for training a tagger for modality
because modality triggers are sparse for the overwhelming majority of sentences.
We investigate an approach to automatically training a modality tagger where we first gathered sentences based on a
high-recall simple rule-based modality tagger and then provided these sentences to
Mechanical Turk annotators for further annotation. 
We used the resulting set of training data to train a precise modality tagger using a multi-class SVM that delivers good performance.
\end{abstract}

\section{Introduction} \label{introduction}

Modality is an extra-propositional component of meaning. In \textit{John
  may go to NY}, the basic proposition is \textit{John go to NY} and the
word {\em may} indicates modality. \newcite{Auwera_2005} define core cases of modality: \textit{John must go to NY} (epistemic necessity), \textit{John might go to NY} 
(epistemic possibility), \textit{John has to leave now} (deontic necessity) and \textit{John may leave now} (deontic possibility). 
Many semanticists (e.g. \newcite{kratzer81}, \newcite{kratzer91}, \newcite{Kaufmann_formalapproaches}) define modality as quantification over possible worlds. 
\textit{John might go} means that there exist some possible worlds in which John goes. 
Another view of modality relates more to a speaker's attitude toward a proposition (e.g. \newcite{McShaneEtAl:2004}).

Modality
might 
be construed broadly to include 
several types of
attitudes that a speaker  
wants to express towards an event, state or proposition.
Modality might indicate
factivity, evidentiality, or sentiment \cite{McShaneEtAl:2004}.  Factivity is related to
whether the speaker wishes to convey his or her belief that the propositional
content is
true or not, i.e., whether it actually obtains in this world or not.  It
distinguishes things that (the speaker believes) happened from things that he or
she desires,
plans, or considers merely probable.  Evidentiality deals with the source of
information and may provide clues to the reliability of the
information.  Did the speaker have firsthand knowledge of what he or
she is reporting, or was it hearsay or inferred from indirect
evidence?  Sentiment deals with a speaker's positive or negative
feelings toward an event, state, or proposition.  

In this paper, we focus on the following five modalities; we have
investigated the belief/factivity modality previously
\cite{diab-EtAl:2009:LAW-III,Prabhakaran:2010}, and we leave other
modalities to future work.
\begin{itemize}
\item {\bf Ability:} can H do P?
\item {\bf Effort:} does H try to do P?
\item {\bf Intention:} does H intend P?
\item {\bf Success:} does H succeed in P?
\item {\bf Want:} does H want P?
\end{itemize}
We investigate
automatically training a modality tagger by using multi-class Support
Vector Machines (SVMs).  
One of the main hurdles for training a linguistic tagger is gathering training data.  
This is particularly problematic for training a modality tagger because modality triggers are sparse for the overwhelming majority of the sentences.
\newcite{baker2010} created a modality tagger by using a semi-automatic
approach for creating rules for a rule-based tagger. 
A pilot study 
revealed that 
it
can boost recall
well above the naturally occurring proportion of modality without annotated
data but with only 60\% precision.  
We investigated an approach
where we first gathered sentences based on a simple modality tagger and
then provided these sentences to annotators for further annotation,
The
resulting annotated data also preserved the level of inter-annotator agreement
for each example so that learning algorithms could take that into account
during training.  Finally, the resulting set of annotations was used for
training a modality tagger using SVMs, which gave a high precision indicating the success of this approach.

Section~\ref{related} discusses related work. 
Section~\ref{data} discusses our procedure for gathering training data. Section~\ref{learner} discusses the
machine learning setup and features used to train our modality tagger and presents experiments and results. Section~\ref{conclusions} concludes and discusses future
work.    

\section{Related Work} \label{related}

Previous related work includes TimeML \cite{SauriVP06}, which involves
modality annotation on events, and Factbank \cite{SauriP09}, where event
mentions are marked with degree of factuality.  Modality is also important
in the detection of uncertainty and hedging.  The CoNLL shared task in
2010~\cite{CoNLL-ST:2010} deals with automatic detection of uncertainty and
hedging in Wikipedia and biomedical sentences.

\newcite{baker2010} and \newcite{baker-et-al:2012} analyze a set of eight
modalities which include belief, require and permit, in addition to the
five modalities we focus on in this paper.  They built a rule-based
modality tagger using a semi-automatic approach to create rules.  This
earlier work differs from the work described in this paper in that the our
emphasis is on the creation of an {\it automatic\/} modality tagger using
machine learning techniques.  Note that the annotation and automatic
tagging of the belief modality (i.e., factivity) is described in more
detail in \cite{diab-EtAl:2009:LAW-III,Prabhakaran:2010}.

There has been a considerable amount of interest in modality in the
biomedical domain.  Negation, uncertainty, and hedging are annotated in the
Bioscope corpus~\cite{Vincze-EtAl:2008}, along with information about which
words are in the scope of negation/uncertainty.  The i2b2 NLP Shared Task
in 2010 included a track for detecting assertion status (e.g. present,
absent, possible, conditional, hypothetical etc.) of medical problems in
clinical records.\footnote{https://www.i2b2.org/NLP/Relations/}
\newcite{Apostolova_2011} presents a rule-based system for the detection of
negation and speculation scopes using the Bioscope corpus.  Other studies
emphasize the importance of detecting uncertainty in medical text
summarization~\cite{morante-daelemans:2009:BioNLP,aramaki-EtAl:2009:BioNLP}.

Modality has also received some attention in the context of certain
applications.  Earlier work describing the difficulty of correctly
translating modality using machine translation includes~\cite{Sigurd:1994}
and~\cite{Murata:2005}.  Sigurd et al.~\shortcite{Sigurd:1994} write about
rule based frameworks and how using alternate grammatical constructions
such as the passive can improve the rendering of the modal in the target
language.  Murata et al.~\shortcite{Murata:2005} analyze the translation of
Japanese into English by several systems, showing they often render the
present incorrectly as the progressive.  The authors trained a support
vector machine to specifically handle modal constructions, while our modal
annotation approach is a part of a full translation system.

The textual entailment literature includes modality annotation schemes.
Identifying modalities is important to determine whether a text entails a
hypothesis.  Bar-Haim et al. \shortcite{Bar-Haim:2007} include polarity
based rules and negation and modality annotation rules.  The polarity rules
are based on an independent polarity lexicon \cite{Nairn:2006}.  The
annotation rules for negation and modality of predicates are based on
identifying modal verbs, as well as conditional sentences and modal
adverbials.  The authors read the modality off parse trees directly using
simple structural rules for modifiers.

\section{Constructing Modality Training Data} \label{data}

In this section, we will discuss the procedure we followed to construct the training data for building the automatic modality tagger.
In a pilot study, we obtained and ran the modality tagger described in
\cite{baker2010} on the English side of the Urdu-English LDC language
pack.\footnote{LDC Catalog No.: LDC2006E110.}  We randomly selected 1997
sentences that the tagger had labeled as not having the Want modality and
posted them on Amazon Mechanical Turk (MTurk).
Three different Turkers (MTurk annotators) marked, for each of the sentences, whether it contained the Want modality.
Using majority rules as the Turker judgment, 95 (i.e., 4.76\%) of these
sentences were marked as having a Want modality.  We also posted 1993
sentences that the tagger had labeled as having a Want modality and only 1238 of
them were marked by the Turkers as having a Want modality.  Therefore, the
estimated precision of this type of approach is only around 60\%.

Hence, we will not be able to use the \cite{baker2010} tagger to gather training data. 
Instead, 
our approach
was to apply a simple tagger as a first
pass, with positive examples subsequently hand-annotated using MTurk.
We made use of sentence data from the Enron email
corpus,\footnote{http://www-2.cs.cmu.edu/$\sim$enron/} derived from the version owing
to Fiore and
Heer,\footnote{http://bailando.sims.berkeley.edu/enron/enron.sql.gz} further
processed as described by \cite{roark2009}.\footnote{Data received through personal communication}
  
To construct the simple tagger (the first pass), we used 
a lexicon of modality trigger words 
(e.g., {\it try, plan, aim, wish, want}) 
constructed by \newcite{baker2010}.  The tagger essentially tags each
sentence that has a word in the lexicon with the corresponding modality.
We wrote a few simple obvious filters for a handful of exceptional cases
that arise due to the fact that our sentences are from e-mail. For example,
we filtered out {\em best wishes}
expressions, which otherwise would have been tagged as {\em Want} because of the
word {\em wishes}.

The words that trigger modality occur with very different frequencies. 
If one is not careful, the training data may be dominated by only the commonly occurring trigger words and the learned tagger would then be biased towards these words.
In order to ensure that our training data had a diverse set of examples
containing many lexical triggers and not just a lot of examples with the
same lexical trigger, for each modality we capped the number of sentences
from a single trigger to be at most 50. 
After we had the set of sentences
selected by the simple tagger, we posted them on MTurk for annotation.  

The
Turkers were asked to check a box indicating that the
modality was not present in the sentence if the 
given modality was not expressed.
If they did not check that box, then they were asked to
highlight the target of the modality.  Table~\ref{t:counts} shows the
number of sentences we posted on MTurk for each modality.\footnote{More detailed statistics on MTurk annotations are available at 
http://hltcoe.jhu.edu/datasets/.}
Three Turkers annotated each sentence. We restricted the task to Turkers who were
adults, had greater than a 95\% approval rating, and had completed at least
50 HITs (Human Intelligence Tasks) on MTurk. We paid US\$0.10 for each set
of ten sentences.

\begin{table}
\centering
\setlength{\extrarowheight}{3pt}
\begin{center}
\begin{tabular}{|l|c|} \hline
Modality    & Count \\ \hline
\hline
Ability     & 190 \\ 
Effort      & 1350 \\
Intention   & 1320 \\
Success     & 1160 \\ 
Want        & 1390 \\ 
\hline
\end{tabular}
\caption{\label{t:counts} For each modality, the number of sentences returned by the simple tagger that we posted on MTurk.} 
\end{center}
\end{table}

Since our data was annotated by three
Turkers, for training data we used only those examples for which at least two
Turkers agreed on the modality and the target of the modality. This
resulted in 
1,008 examples.
674 examples had two Turkers agreeing and 334
had unanimous agreement. We kept track of the level of agreement for each
example so that our learner could weight the examples differently depending
on the level of inter-annotator agreement. 

\section{Multiclass SVM for Modality} \label{learner}

In this section, we describe the automatic modality tagger we built using the MTurk annotations described in Section \ref{data} as the training data. 
Section \ref{sec:data} describes the training and evaluation data. 
In Section \ref{sec:approach}, we present the 
machinery and Section \ref{sec:feats} describes the features we used to train the tagger. 
In Section \ref{sec:exp}, we present various experiments and discuss results. 
Section \ref{sec:confexp}, presents additional experiments using annotator confidence.

\subsection{Data}
\label{sec:data}

For training, we used the data presented in Section \ref{data}. 
We refer to it as MTurk data in the rest of this paper.
For evaluation, we selected a part of the LU Corpus \cite{LU_Corpus}
(1228 sentences) and our expert annotated it with modality tags.
We first used
the high-recall simple modality tagger described in Section \ref{data} to
select the sentences with modalities. Out of the 235 sentences returned by
the simple modality tagger, our expert removed the ones which did not in
fact have a modality.  In the remaining sentences (94 sentences), our
expert annotated the target predicate. 
We refer to this as the Gold dataset in this paper. The MTurk and Gold datasets differ in terms of genres as well as annotators (Turker vs. Expert).
The distribution of modalities in both MTurk and Gold
annotations are given in Table \ref{freq}.

\begin{table}[h]
\centering
\setlength{\extrarowheight}{3pt}
\begin{tabular} { | l | c | c | }
\hline
Modality & MTurk  & Gold  \\
\hline
\hline
Ability & 6\% & 48\% \\
Effort & 25\% & 10\% \\
Intention & 30\% & 11\% \\
Success & 24\% & 9\% \\
Want & 15\% & 23\% \\
\hline
\end{tabular}
\caption{Frequency of Modalities}
\label{freq}
\end{table}

\subsection{Approach}
\label{sec:approach}

We applied a supervised learning framework using multi-class SVMs 
to automatically learn to tag modalities in context. For tagging, we used the Yamcha
\cite{kudo/matsumoto:2003a} sequence labeling system which uses the 
SVM$^{\rm light}$ \cite{Joachims:1999}
package for classification. We used \textit{One versus All} method for multi-class classification
on a quadratic kernel with a C value of 1.
We report recall and precision on word tokens in our corpus for each
modality.  We also report F$_{\beta=1}$ (F)-measure as the harmonic mean
between (P)recision and (R)ecall.

\subsection{Features}
\label{sec:feats}

We used lexical features at the token level which can be
extracted without any parsing with relatively high accuracy. We use the
term context width to denote the window of tokens whose features are
considered for predicting the tag for a given token. For example, a context
width of 2 means that the feature vector of any given token includes, in
addition to its own features, those of 2 tokens before and after it as well
as the tag prediction for 2 tokens before it.
We did experiments varying the context width from $1$ to $5$ and found that
a context width of $2$ gives the optimal performance. All results reported
in this paper are obtained with a context width of $2$.
For each token, we performed experiments using following lexical features:
\begin{itemize}
\item \textbf{wordStem} - Word stem.
\item \textbf{wordLemma} - Word lemma.
\item \textbf{POS} - Word's POS tag.
\item \textbf{isNumeric} - Word is Numeric?
\item \textbf{verbType} - Modal/Auxiliary/Regular/Nil 
\item \textbf{whichModal} - If the word is a modal verb, which modal? 
\end{itemize}

We used the Porter stemmer \cite{Porter:1997} to obtain the stem of a word
token. To determine the word lemma, we used an in-house lemmatizer using
dictionary and morphological analysis to obtain the dictionary form of a
word. We obtained POS tags from Stanford POS tagger and used those tags to
determine \textit{verbType} and \textit{whichModal} features. The
\textit{verbType} feature is assigned a value `Nil' if the word is not a
verb and \textit{whichModal} feature is assigned a value `Nil' if the word
is not a modal verb. The feature \textit{isNumeric} is a binary feature
denoting whether the token contains only digits or not.

\subsection{Experiments and Results}
\label{sec:exp}

In this section, we present experiments performed considering                    
all the MTurk annotations where two annotators agreed and all the MTurk annotations where all three annotators agreed to be equally correct annotations. 
We present experiments applying differential weights for these annotations in Section \ref{sec:confexp}.
We performed 4-fold cross validation (4FCV) on MTurk data in order to select
the best feature set configuration $\phi$. 
The best feature set obtained
was $wordStem, POS, whichModal$ with a context width of $2$.
For finding the best performing feature set - context width configuration,
we did an exhaustive search on the feature space, pruning away features which
were proven not useful by results at stages.
Table \ref{tab:best4fcv} presents results obtained for each modality on 4-fold cross validation.

\begin{table}[ht]
\centering
\setlength{\extrarowheight}{3pt}
\begin{tabular} { | l | c | c | c | }
\hline
Modality & Precision & Recall & F Measure \\
\hline
\hline
Ability & 82.4 & 55.5 & 65.5\\
Effort & 95.1 & 82.8 & 88.5 \\
Intention & 84.3 & 61.3 & 70.7 \\
Success & 93.2 & 76.6 & 83.8 \\
Want & 88.4 & 64.3 & 74.3 \\
\hline
Overall & \textbf{90.1} & \textbf{70.6} & \textbf{79.1} \\
\hline
\end{tabular}
\caption{Per modality results for best feature set $\phi$ on 4-fold cross validation on MTurk data}
\label{tab:best4fcv}
\end{table}

We also trained a model on the entire MTurk data using the best feature
set $\phi$ and evaluated it against the Gold data. The results obtained for
each modality on gold evaluation are
given in Table \ref{tab:bestgold}.  We attribute the lower performance on
the Gold dataset to its difference from MTurk data. MTurk data is entirely
from email threads, whereas Gold data contained sentences from newswire,
letters and blogs in addition to emails.  Furthermore, the annotation is
different (Turkers vs expert).  Finally, the distribution of modalities in
both datasets is very different. For example, {\em Ability} modality was merely 6\%
of MTurk data compared to 48\% in Gold data (see Table~\ref{freq}). 

\begin{table}[ht]
\centering
\setlength{\extrarowheight}{3pt}
\begin{tabular} { | l | c | c | c | }
\hline
Modality & Precision & Recall & F Measure \\
\hline
\hline
Ability & 78.6 & 22.0 & 34.4\\
Effort & 85.7 & 60.0 & 70.6\\
Intention & 66.7 & 16.7 & 26.7\\
Success & NA & 0.0 & NA\\
Want & 92.3 & 50.0 & 64.9\\
\hline
Overall & \textbf{72.1} & \textbf{29.5} & \textbf{41.9}\\
\hline
\end{tabular}
\caption{Per modality results for best feature set $\phi$ evaluated on Gold dataset}
\label{tab:bestgold}
\end{table}

We obtained reasonable performances for \textit{Effort} and \textit{Want} modalities while the performance for other modalities was rather low.
Also, the Gold dataset contained only 8 instances of \textit{Success}, none of which was recognized by the tagger resulting in a recall of 0\%. 
Precision (and, accordingly, F Measure) for \textit{Success} was considered ``not applicable'' (NA), as no such tag was assigned.
                   
\subsection{Annotation Confidence Experiments}
\label{sec:confexp}

\begin{table*}
\setlength{\extrarowheight}{3pt}
\begin{center}
\begin{tabular*} {0.809\textwidth} { | l | c | c | c | c | c | c |}
\hline
\multirow {2}{*}{TrainingSetup} & \multicolumn {3} {c|} {Tested on $Agr_2$ and $Agr_3$} & \multicolumn {3} {c|} {Tested on $Agr_3$ only} \\
\cline{2-7}
& Precision & Recall & F Measure & Precision & Recall & F Measure \\
\hline
\hline
$Tr23$ & 90.1 & {\bf 70.6} & {\bf 79.1} & 95.9 & {\bf 86.8} & {\bf 91.1} \\
$Tr2$ & {\bf 91.0} & 66.1 & 76.5 & 95.6 & 81.8 & 88.2 \\
$Tr3$ & 88.1 & 52.3 & 65.6 & {\bf 96.8} & 71.7 & 82.3 \\
$Tr23_W$ & 89.9 & 70.5 & 79.0 & 95.8 & 86.5 & 90.9 \\
\hline
\end{tabular*}
\caption{Annotator Confidence Experiment Results; the best results per
  column are boldfaced\label{tab:conf:4fcv}}
 (4-fold cross validation on MTurk Data)
 \end{center}
\end{table*}

Our MTurk data contains sentence for which at least two of the three Turkers agreed on the modality and the target of the modality.
In this section, we
investigate the role of annotation confidence in training an automatic tagger.
The annotation confidence is denoted by whether an annotation was agreed by only two annotators or was unanimous.
We denote the set of sentences for which only
two annotators agreed as $Agr_2$ and that for which all three annotators agreed as $Agr_3$.

We present four training setups.
The first setup is $Tr23$ where we train a model using both $Agr_2$ and
$Agr_3$ with equal weights. This is the setup we used for results presented in the Section \ref{sec:exp}. Then, we have $Tr2$ and $Tr3$, where we train
using only $Agr_2$ and $Agr_3$ respectively. Then, for $Tr23_W$, we train a
model giving different cost values for $Agr_2$ and $Agr_3$ examples. 
The SVMLight package allows users to input cost values $c_i$ for each training instance separately.\footnote{This can be done by 
specifying `cost:$<$value$>$' after the label in each training instance. This feature has not yet been documented on the SVMlight website.} 
We tuned this cost value for $Agr_2$ and $Agr_3$ examples and found the best
value at $20$ and $30$ respectively. 

For all four setups,
we used feature set $\phi$.
We performed $4$-fold cross validation on MTurk data in two ways --- we
tested against a combination of $Agr_2$ and $Agr_3$, and we tested against only $Agr_3$. Results of these experiments are presented in 
Table \ref{tab:conf:4fcv}. We also present the results of evaluating a tagger trained on the whole MTurk data for each setup against the Gold annotation in 
Table \ref{tab:conf:gold}.
The $Tr23$ tested on both $Agr_2$ and $Agr_3$ presented in Table \ref{tab:conf:4fcv} and $Tr23$ tested on Gold data presented in Table \ref{tab:conf:gold} correspond 
to the results presented in Table \ref{tab:best4fcv} and Table \ref{tab:bestgold} respectively.

\begin{table}[h]
\centering
\setlength{\extrarowheight}{3pt}
\begin{center}
\begin{tabular} { | l | c | c | c | }
\hline
TrainingSetup & Precision & Recall & F Measure \\
\hline
\hline
$Tr23$ & 72.1 & 29.5 & 41.9 \\
$Tr2$ & 67.4 & 27.6 & 39.2 \\
$Tr3$ & {\bf 74.1} & 19.1 & 30.3 \\
$Tr23_W$ & 73.3 & {\bf 31.4} & {\bf 44.0} \\
\hline
\end{tabular}
\end{center}
\caption{Annotator Confidence Experiment Results; the best results per column are boldfaced\label{tab:conf:gold}}
(Evaluation against Gold)
\end{table}

One main observation is that including annotations of lower agreement, but
still above a threshold (in our case, 66.7\%), is definitely
helpful. $Tr23$ outperformed both $Tr2$ and $Tr3$ in both 
recall and F-measure in all evaluations. Also, even when evaluating against only the high
confident $Agr_3$ cases, $Tr2$ gave a high gain in recall (10 .1 percentage
points) over $Tr3$, with only a 1.2 percentage point loss on precision. We
conjecture that this is because there are far more training instances in
$Tr2$ than in $Tr3$ (674 vs 334), and that quantity beats quality.

Another important observation is the increase in performance by using
varied costs for $Agr_2$ and $Agr_3$ examples (the $Tr23_W$ condition). Although it dropped the
performance by 0.1 to 0.2 points in cross-validation F measure on the Enron
corpora, it gained 2.1 points in Gold evaluation F measure. 
These results seem to indicate that differential weighting based on
annotator agreement might have more beneficial impact when training a model
that will be applied to a wide range of genres than when training a model
with genre-specific data for application to data from the same genre.  Put
differently, using varied costs prevents genre over-fitting.
We don't have a full explanation for this difference in behavior yet. We plan to explore this in future work.

\section{Conclusion} \label{conclusions}

We have presented an innovative way of combining a high-recall simple tagger with Mechanical Turk annotations to produce training data
for a modality tagger.  We show that we obtain good performance on the same
genre as this training corpus (annotated in the same manner), and
reasonable performance across genres (annotated by an independent
expert). 
We also present experiments utilizing the number of agreeing Turkers to choose cost values for training examples for the SVM. 
As future work, we plan to extend this approach to other modalities which are not covered in this study.

\section{Acknowledgments}

This work is supported, in part, by the Johns Hopkins Human Language
Technology Center of Excellence. Any opinions, findings, and conclusions or
recommendations expressed in this material are those of the authors and do
not necessarily reflect the views of the sponsor. 
We thank several anonymous reviewers for their constructive feedback.

\bibliographystyle{acl2012}
\bibliography{paper}

\end{document}